\documentclass[10pt,twocolumn,letterpaper]{article}

\usepackage{btas}
\usepackage{times}
\usepackage{epsfig}
\usepackage{graphicx}
\usepackage{amsmath}
\usepackage{amssymb}
\usepackage[normalem]{ulem}
\usepackage{gensymb}
\usepackage{url}

\usepackage{subcaption}
\usepackage{flushend}

\btasfinalcopy 

\ifbtasfinal\fi
\begin{document}

\title{Open Source Presentation Attack Detection Baseline for Iris Recognition}

\author{Joseph McGrath \hspace{1cm} Kevin W. Bowyer \hspace{1cm} Adam Czajka \\
Department of Computer Science and Engineering, University of Notre Dame\\
{\tt\small \{jmcgrat3,kwb,aczajka\}@nd.edu}
}

\maketitle
\ifbtasfinal\thispagestyle{empty}\fi

\begin{abstract}
This paper proposes the first, known to us, open source presentation attack detection (PAD) solution to distinguish between authentic iris images (possibly wearing clear contact lenses) and irises with textured contact lenses. This software can serve as a baseline in various PAD evaluations, and also as an open-source platform with an up-to-date reference method for iris PAD. The software is written in C++ and Python and uses only open source resources, such as OpenCV. This method does not incorporate iris image segmentation, which may be problematic for unknown fake samples. Instead, it makes a best guess to localize the rough position of the iris. The PAD-related features are extracted with the Binary Statistical Image Features (BSIF), which are classified by an ensemble of classifiers incorporating support vector machine, random forest and multilayer perceptron. The models attached to the current software have been trained with the NDCLD'15 database and evaluated on the independent datasets included in the LivDet-Iris 2017 competition. The software implements the functionality of retraining the classifiers with any database of authentic and attack images. The accuracy of the current version offered with this paper exceeds 99\% when tested on subject-disjoint subsets of NDCLD'15, and oscillates around 85\% when tested on the LivDet-Iris 2017 benchmarks, which is on par with the results obtained by the LivDet-Iris 2017 winner.

\end{abstract}

\section{Introduction}
\label{sec:intro}

Presentation attack detection (PAD) is an important element of biometric systems. There were multiple demonstrations of successful presentation attacks on commercial systems suggesting that the PAD mechanisms were either ineffective or missing in these systems, and iris recognition is not an exception. Starting from the first experiments in 2002 showing that paper printouts can be matched to real irises \cite{Thalheim_CT_2002} by a commercial system, and running through spoofing of iris recognition in the Samsung Galaxy S8\footnote{\url{https://goo.gl/zjEF3M}, The Guardian, May 2017}, we can conclude that iris PAD is not a solved problem. The most recent LivDet-Iris 2017 evaluation \cite{Yambay_IJCB_2017} additionally suggests that the open-set regime, in which some (or all) properties of samples are unknown during training, is even more challenging, as the winning algorithm did not recognize from 11\% to 38\% of attack images, depending on the database.

Iris PAD is a dynamic research area, with many algorithms proposed to date \cite{Czajka_CSUR_2018}. The question arises: which factors prevent us, as a community, from moving forward with making iris PAD more effective, especially for unknown attack types? One possible reason is the lack of an open source platform to maintain a baseline iris PAD methodology that is easy to contribute to and easy to benefit from when developing or evaluating original solutions. The OpenCV platform\footnote{\url{https://opencv.org}} is a great example of such an initiative in computer vision in general. The Masek's implementation \cite{Masek_2003} and more recently the OSIRIS system \cite{osiris} have played this role for iris recognition. This paper 
offers the first open source software based on a strong, recent methodology of textured contact lens detection employing Binary Statistical Image Features (BSIF) and ensemble classification realized by Support Vector Machines (SVM), Random Forests (RF), and multilayer perceptrons (MLP) \cite{Doyle_IEEEAccess_2015}. The initial version proposed in this paper includes the classifier ensemble already trained on one of the publicly available datasets of images of irises with and without textured contact lenses, NDCLD'15\footnote{\url{https://cvrl.nd.edu/projects/data/\#the-notre-dame-contact-lense-dataset-2015ndcld15}}, and hence is ``ready to use''. However, one of the functionalities of this software is to retrain the ensemble with any samples, especially those which conform to the ISO/IEC 19794-6 standard. The proposed solution also delivers raw BSIF-based PAD features for those who want to test other classifiers and ensembles. It is specialized (in this current version) to detection of textured contact lenses, however re-training for other presentation attack instruments (for instance, paper printouts) is straightforward. The GitHub repository with the software and ready to use models can be accessed at \url{https://github.com/CVRL/OpenSourceIrisPAD}
. To our knowledge, this is the first and only proposal of an open source solution for iris presentation attack detection.


\section{Related Work}
\label{sec:related}

The number of iris PAD methods developed to date is significant, and a recent survey by Czajka and Bowyer \cite{Czajka_CSUR_2018} categorizes them into groups of methods using either still iris images or iris videos, which are either acquired passively (with no eye stimuli) or actively (when the eye is stimulated by external light, or a response is expected from the subject). The group of methods using still samples in PAD, identical to those used in iris recognition, is mostly populated by solutions employing various texture descriptors (such as BSIF \cite{Komulainen_IETbook_Ch12_2017}), or -- recently popular -- convolutional neural networks \cite{Chen_WACV_2018}. If some modifications in the iris recognition equipment are possible, the iris PAD methods incorporate multi-spectral imaging solely in near-infrared band \cite{Park_OptEng_2007} or combined with visible-light imaging \cite{Thavalengal_TCE_2016}, 3D properties of the eye \cite{Lee_IMA_2010,Czajka_WACV_2019}, or dynamic features such as spontaneous \cite{Villalbos-Castaldi_BF_2014} or stimulated \cite{Czajka_TIFS_2015} pupil oscillations, eye blinks \cite{Raja_TIFS_2015}, or eyeball movements \cite{Komogortsev_TIFS_2015}. Despite the large number of proposed PAD methods to date, Czajka and Bowyer \cite{Czajka_CSUR_2018} conclude that they ``do not know of even a single well-documented iris PAD algorithm that is available to the research community as open source,'' which motivated our publishing of this first open-source solution.

In the context of the existing tools for and efforts towards faster development of iris PAD methodologies, it is worth mentioning numerous benchmark databases, such as {\sf Clarkson}, {\sf Warsaw}, {\sf Notre Dame}, and {\sf WVU/IIITD-Delhi} developed for LivDet-Iris competitions \cite{Yambay_IJCB_2014,Yambay_ISBA_2017,Yambay_IJCB_2017} (paper printouts and textured contact lenses), {\sf NDCCL 2012} \cite{Doyle_ICB_2013}, {\sf NDCLD 2013}, \cite{Doyle_BTAS_2013} and {\sf NDCLD 2015} \cite{Doyle_IEEEAccess_2015} (clear and textured contact lenses), {\sf ATVS-FIr} \cite{Galbally_ICB_2012} (paper printouts), {\sf Pupil-Dynamics} \cite{Czajka_TIFS_2015} (pupil size in time with and without visible-light stimuli), {\sf Post-Mortem-Iris} \cite{Trokielewicz_BTAS_2016} (images of irises acquired up to one month after death), {\sf CASIA-Iris-Syn} \cite{Wei_ICPR_2008} (synthetically generated iris images), or data acquired by a LightField sensor {\sf GUC-LF-VIAr-DB} \cite{Raghavendra_IJCB_2014}. The LivDet-Iris competitions\footnote{\url{http://livdet.org}}, mentioned earlier, are an important effort towards independent evaluation of iris PAD algorithms. Editions were organized in 2013 \cite{Yambay_IJCB_2014}, 2015 \cite{Yambay_ISBA_2017}, and 2017 \cite{Yambay_IJCB_2017} and brought together researchers from around the world, who submitted their iris PAD algorithms for evaluation.

It is also worth mentioning the ISO PAD-related standardization efforts. In particular, the ISO/IEC 30107-1 standard defines the PAD framework and vocabulary, and is freely available\footnote{\url{https://goo.gl/JSbiqy}}. The third part, ISO/IEC 30107-3 defining the PAD evaluation, has been adopted to the National Voluntary Laboratory Accreditation Program (NVLAP) run by NIST\footnote{\url{https://www.nist.gov/nvlap}}.

We hope that the proposed open-source iris PAD software will fill the current gap of strong, recent open-source baseline iris PAD algorithms, stimulate their development, and see multiple contributors.

\section{The Baseline Method for Textured Contact Lens Detection}
\label{sec:method}

\begin{figure*}[!htb]
    \centering
    \begin{subfigure}[t]{0.19\textwidth}
        \includegraphics[width=\textwidth]{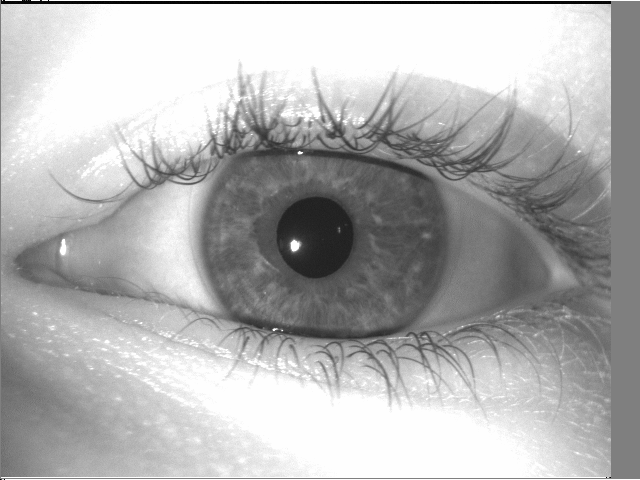}
        \caption{Live iris}
    \end{subfigure}\hfill
    \begin{subfigure}[t]{0.19\textwidth}
        \includegraphics[width=\textwidth]{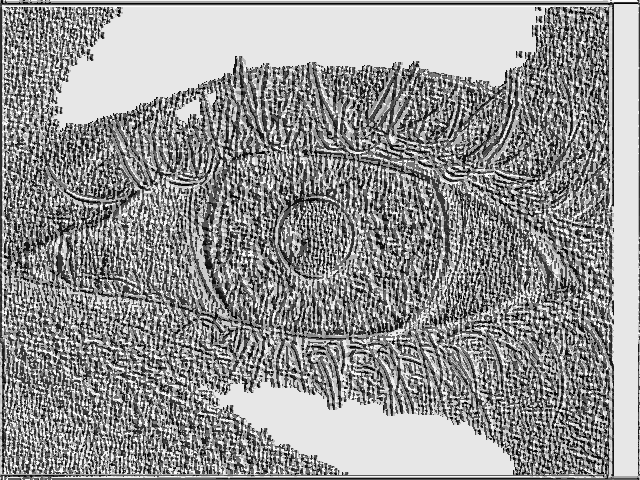}
        \caption{$7\times7$ code of (a)}
    \end{subfigure}\hfill
    \begin{subfigure}[t]{0.19\textwidth}
        \includegraphics[width=\textwidth]{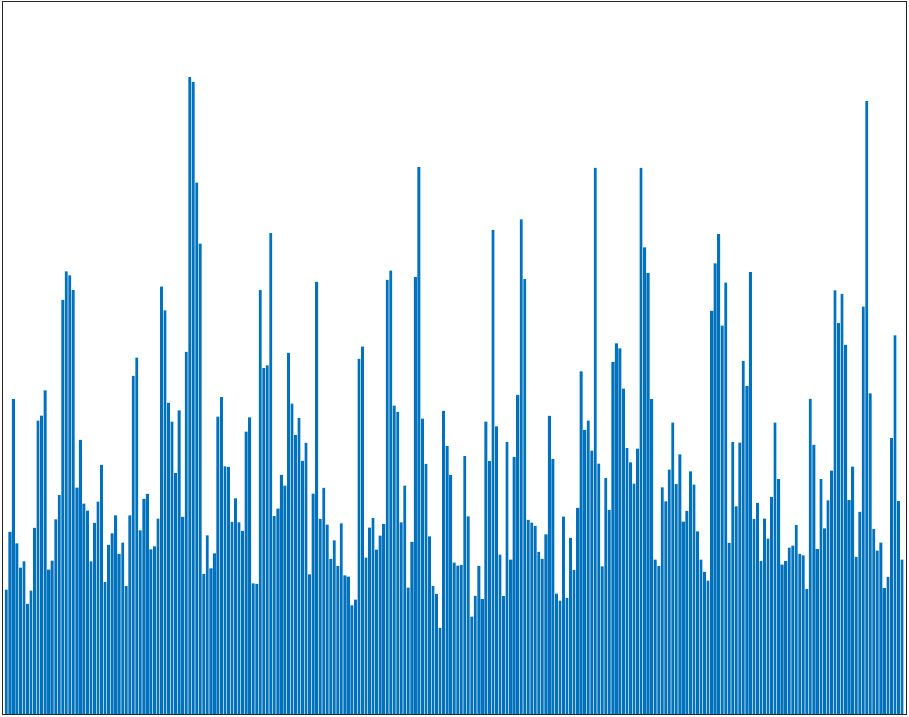}
        \caption{Histogram of (b)}
    \end{subfigure}\hfill
    \begin{subfigure}[t]{0.19\textwidth}
        \includegraphics[width=\textwidth]{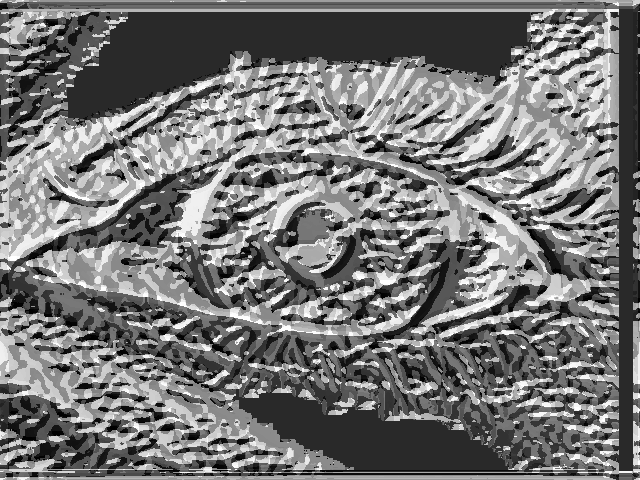}
        \caption{$17\times17$ code of (a)}
    \end{subfigure}\hfill
    \begin{subfigure}[t]{0.19\textwidth}
        \includegraphics[width=\textwidth]{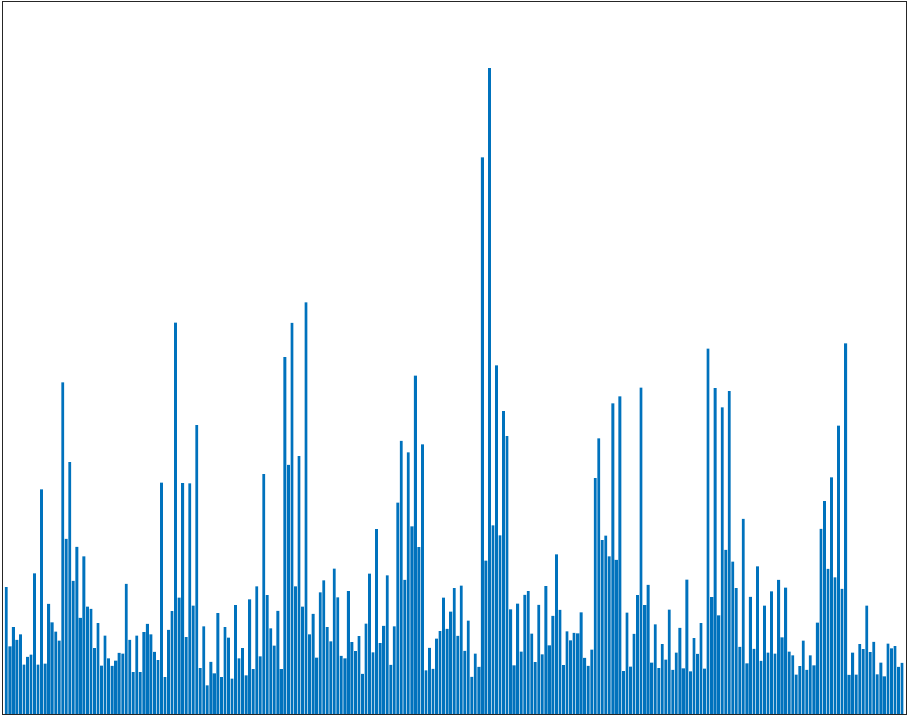}
        \caption{Histogram of (d)}
    \end{subfigure}\\
    \begin{subfigure}[t]{0.19\textwidth}
        \includegraphics[width=\textwidth]{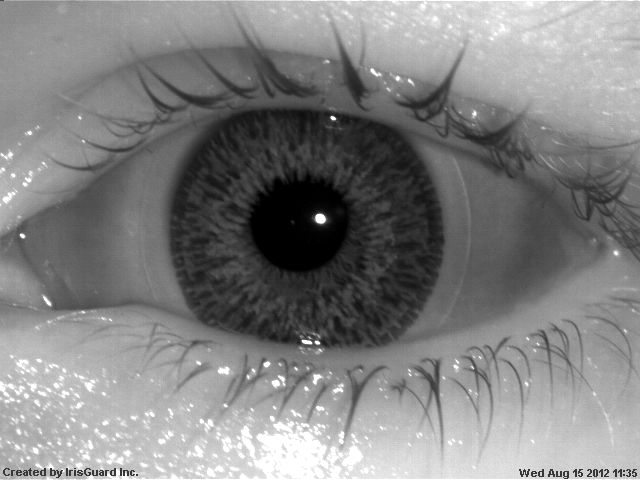}
        \caption{Textured contact}
    \end{subfigure}\hfill
    \begin{subfigure}[t]{0.19\textwidth}
        \includegraphics[width=\textwidth]{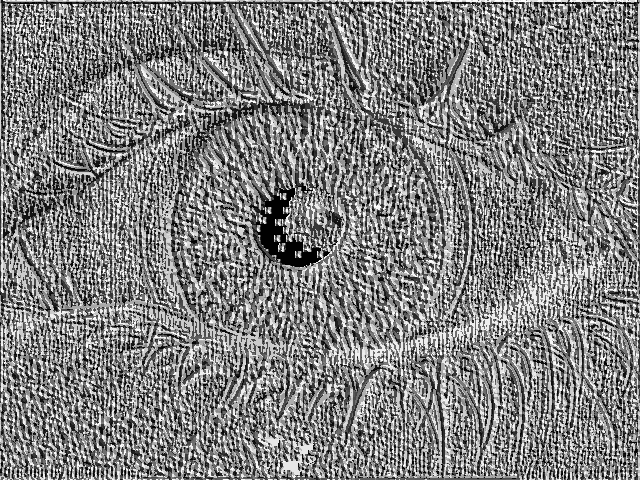}
        \caption{$7\times7$ code of (f)}
    \end{subfigure}\hfill
    \begin{subfigure}[t]{0.19\textwidth}
        \includegraphics[width=\textwidth]{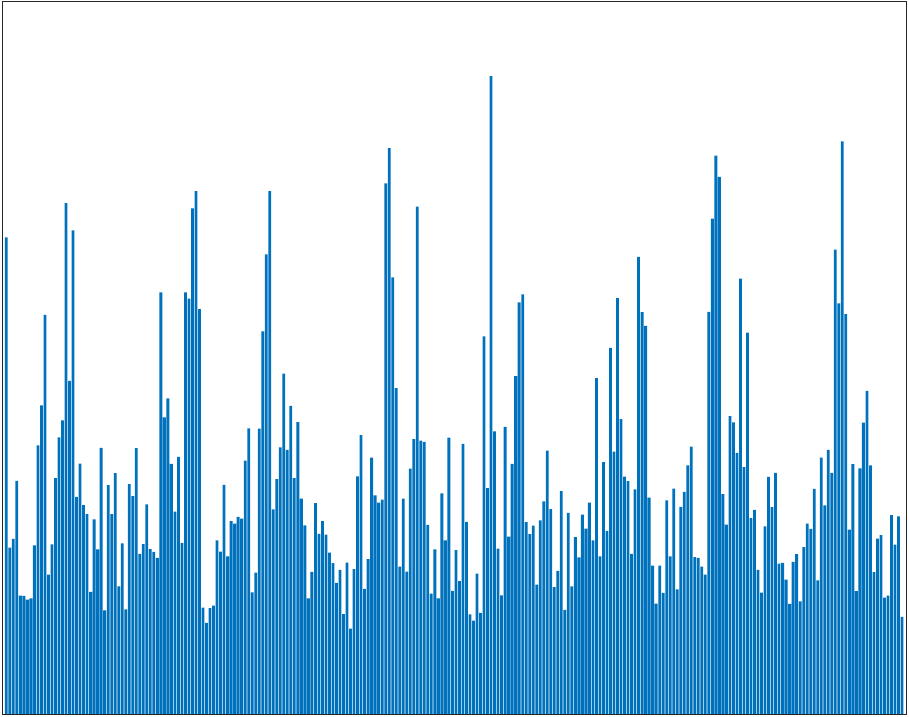}
        \caption{Histogram of (g)}
    \end{subfigure}\hfill
    \begin{subfigure}[t]{0.19\textwidth}
        \includegraphics[width=\textwidth]{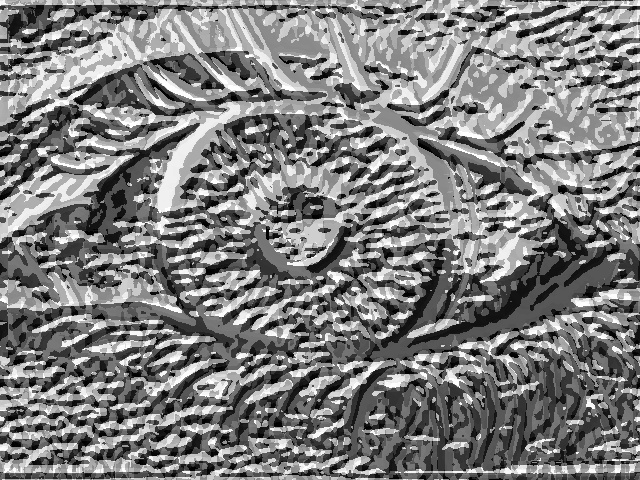}
        \caption{$17\times17$ code of (f)}
    \end{subfigure}\hfill
    \begin{subfigure}[t]{0.19\textwidth}
        \includegraphics[width=\textwidth]{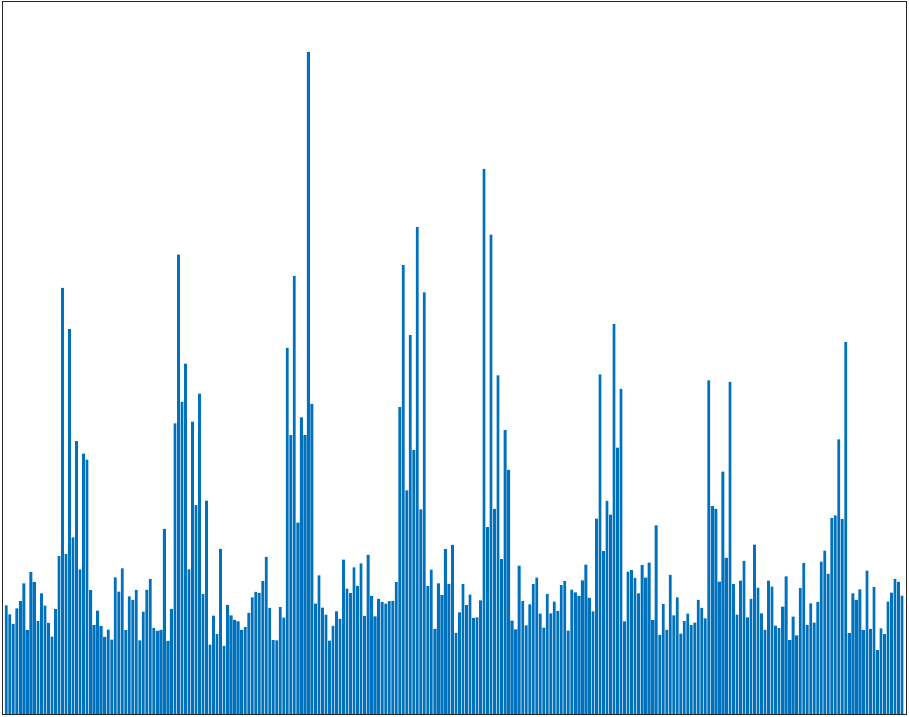}
        \caption{Histogram of (i)}
    \end{subfigure}
    \caption{8-bit BSIF codes and the resulting histograms calculated at two example scales for an authentic iris image and an iris image with textured contact lens.}
    \label{fig:BSIFexamples}
\end{figure*}

The implemented solution extends the methodology proposed by Doyle and Bowyer \cite{Doyle_IEEEAccess_2015} and the feature extraction is based on Binary Statistical Image Features (BSIF) proposed by Kannala and Rahtu \cite{Kannala_ICPR_2012}. In this method, the calculated ``BSIF code'' is based on filtering the image with $n$ filters of size $s\times s$, and then binarizing the filtering results with a threshold at zero. Hence, for each pixel $n$ binary responses are given, which are in the next step translated into a $n$-bit grayscale value. In the original BSIF paper, $n \in \{5, 6, 7, 8, 9, 10, 11, 12\}$, and $s \in \{3, 5, 7, 9, 11, 13, 15, 17\}$, and thus there are 60 combinations of $n$ and $s$ (4 combinations, namely $n \in \{9, 10, 11, 12\}$ for $s=3$, were skipped). The filters, for each considered combination of $n$ and $s$, were trained on patches extracted from natural images in a way to maximize the statistical independence of filter responses. Fig. \ref{fig:BSIFexamples} presents BSIF codes for example iris images (with and without textured contact lenses) for two example scales ($s=7$ and $s=17$) and $n=8$.

The histograms resulting from gray-scale BSIF codes are later normalized to a $z$-score and used as texture descriptors with the number of histogram bins equal to $2^n$, as shown in Fig. \ref{fig:BSIFexamples}. Following \cite{Doyle_IEEEAccess_2015}, we extract BSIF codes for an image down-sampled to $320\times240$ in addition to the original ISO-complaint iris image resolution of $640\times480$. This allows for the exploration of more scales in feature extraction. The \textit{Best Guess} segmentation technique \cite{Doyle_IEEEAccess_2015} has been implemented as standard in the baseline method: a region of interest is selected that corresponds to the average iris center point and radius across the training set. For the ISO-compliant iris images in the NDCLD'15 dataset, this corresponds to a center location of $(320, 250)$ with a radius of $125$ pixels. 

To extend the original methodology \cite{Doyle_IEEEAccess_2015}, in which only $n=8$ filters were used, this method has been implemented for all values of $n$, as proposed in the original BSIF paper. This allows for richer feature sets and more options when searching for optimal ensemble of classifiers. Consequently, there are 120 histograms for each image, and a separate set of three classifiers (SVM, RF, and MP) is trained for each feature set, giving 360 ``experts'' ready to vote for each input image. Since not all the classifiers have the same strength, a subset of the strongest classifiers is selected and majority voting is applied to these selected classifiers to come up with a final decision. The set of strongest classifiers can be configured in the proposed solution.

\begin{figure*}[!htb]
    \centering
    \includegraphics[width=\textwidth]{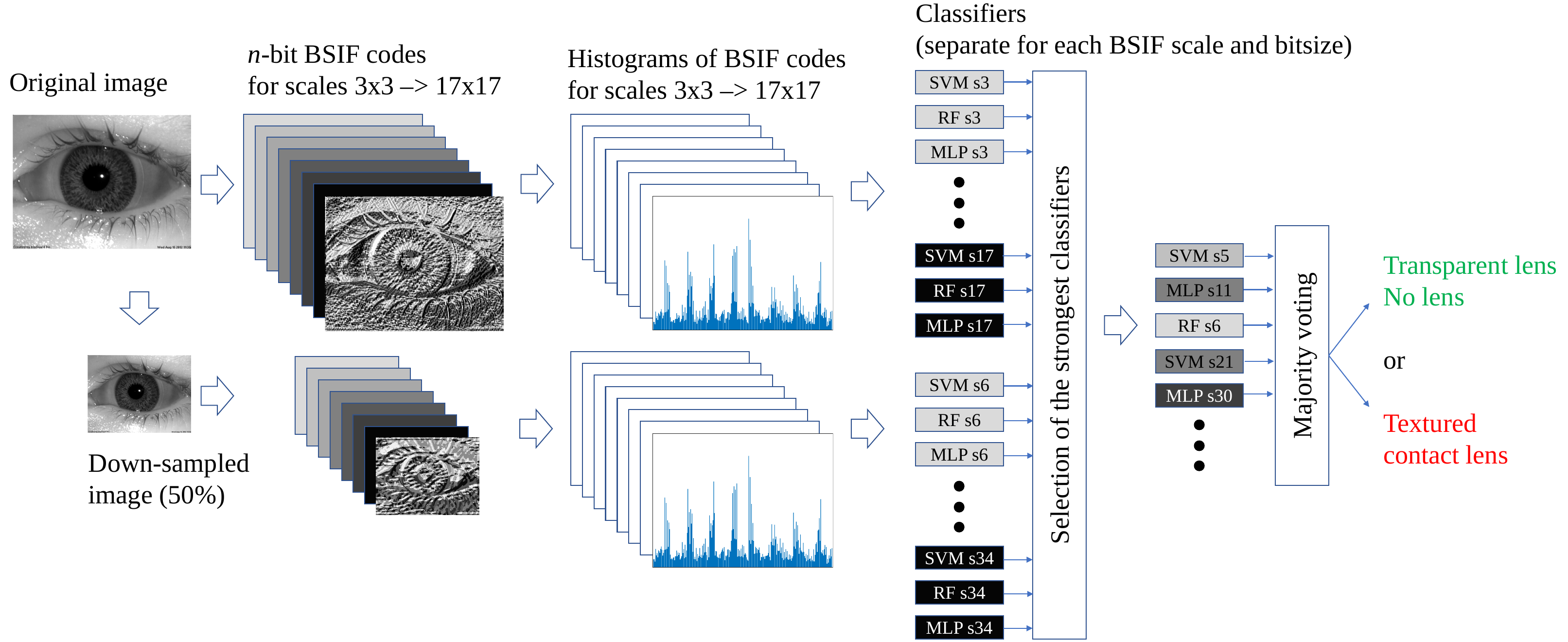}
    \caption{The schematic view of the proposed open-source iris PAD solution.}
    \label{fig:schema}
\end{figure*}

\section{Software Architecture}
\label{sec:architecture}

The \verb+TCL Detection+ solution includes versions written in C++ and Python. It has been tested on MacOS High Sierra 10.13.6 using g++ 4.2.1 as the compiler, and on Windows 7 64 bit using Microsoft Visual Studio 2015. The implementation uses two external libraries: OpenCV version 3.4.1 for computer vision functionality and HDF5 version 1.10.4 for the feature storage file format. \verb+TCL Detection+ is organized into three main modes of operation: feature extraction, model training, and model testing.  

The overall structure of both versions (C++ and Python) of \verb+TCL Detection+ is based on the structure of OSIRIS version 4.1 in that a manager class is used to control all information flow within the application.  The manager contains methods to parse the configuration file, show the configuration, and run the desired mode of operation by creating instances of other classes necessary to the operation of the program.

The \verb+featureExtractor+ class (\verb+filter+ module in Python) handles the extraction of BSIF features when given a list of image files to be used in extraction. The \verb+extract()+ method can be used to extract and save BSIF features for a specific scale $s$ and number of bits $n$. This method creates an instance of the BSIF filter that is then used to filter a regular or down-sampled version of each image, depending on the input scale. The \verb+featureExtractor+ class and \verb+filter+ module also serve as the interface with the required HDF5 functions: they create a new file for each combination of $s,n$ and place the feature sets within the file, indexed by the name of the image they represent.

The \verb+BSIFFilter+ class contains the methods for loading the hard-coded BSIF filters and for generating a histogram for an image using the method described in \ref{sec:method}. In the Python implementation, the C++ version of the \verb+BSIFFilter+ class is used with Python bindings to load the required filters. The \verb+generateHistogram()+ method assigns a bit to each filter used, giving an $n$-bit integer for each pixel corresponding to the response of the filters. A histogram is then taken across the entire image and returned to the \verb+extract+ method, which saves the histogram as the feature set for that image, bit size, and filter scale.

The remainder of the operation modes are handled by the manager class, which instantiates the required OpenCV objects to train models and test images. If model training is enabled, the manager will call a method, \verb+loadFeatures+, to load the features for the images specified in the training list. The training features and classifications are then loaded into an instance of the \verb+TrainData+ class from \verb+OpenCV+ for C++ or a Numpy array for Python. A new model of the type specified in the configuration file is then initialized and trained; currently, the supported model types are SVM, random forest, and multilayer perceptron. Training is achieved through the \verb+trainAuto+ function in OpenCV for SVM and through custom training functions designed to replicate the function of \verb+trainAuto+ for random forest and multilayer perceptron. These training functions choose the optimal parameters using k-fold cross validation with ten folds. Each model that is trained is then output as an \verb+xml+ file.

If model testing is enabled, the manager will load the required models from their \verb+xml+ files. If majority voting is disabled, the manager will individually load each model and the testing features corresponding to the BSIF scale and bit size the model was trained on. These features will then be input to the \verb+predict+ function for the model from OpenCV and the predictions will be returned. The predictions will be tested against the classifications provided with the test set and the CCR, APCER, and BPCER will be output for each model individually.

If majority voting is enabled, the predictions for each model are determined and temporarily stored. For each image, the number of models voting for each classification is determined and the overall decision is made with a simple majority vote. In the case of a tie, a random decision is made. The ensemble accuracy on the training set is then determined through comparison with the classifications provided with the test set.

\section{Results}
\subsection{Datasets}
NDCLD'15 was used as the primary training dataset for all experiments \cite{Doyle_IEEEAccess_2015}. The 7300 images in this dataset were acquired with two different sensors -- IrisGuard AD100 and IrisAccess LG4000 -- that are equally represented in the dataset. The dataset includes images with no contact lenses, clear contact lenses, and textured contact lenses from five different brands, which are equally represented with 500 images each: J\&J, Ciba, Cooper, UCL, and ClearLab.

Two additional datasets used in the LivDet-Iris 2017 competition, Clarkson and IIITD, were used for the validation and testing of the ensemble \cite{Yambay_IJCB_2017}. The Clarkson dataset was collected using the LG IrisAccess EOU2200 and consists of 2469 live iris images, 1122 textured contact lens images, and printouts of iris images. The printout images were not used, as the focus of this open source baseline is textured contact lenses as a presentation attack instrument. The IIITD dataset consists of 2250 live iris images and 1000 textured contact lens images.

\subsection{Base Case}
\begin{figure*}[!htb]
        \includegraphics[width=\textwidth, keepaspectratio]{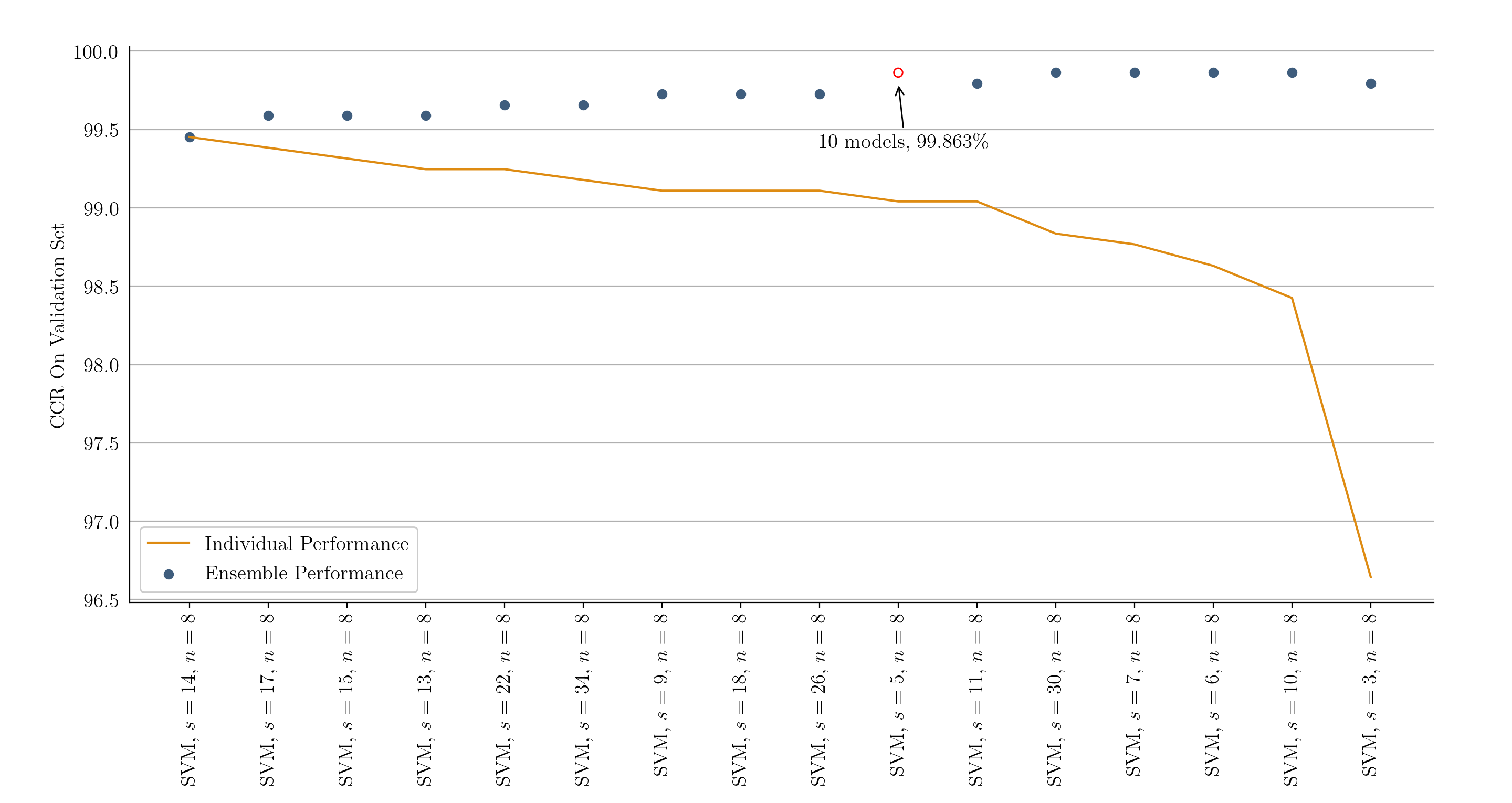}
        \caption{Building a classification ensemble on {\bf Notre Dame} dataset.}
        \label{fig:nd_validation}
\end{figure*}

Following Doyle \textit{et al.} \cite{Doyle_IEEEAccess_2015}, a base case was tested with SVM models trained on BSIF with $n=8$ and all scales mentioned in \ref{sec:method}, giving 16 total models. SVM was selected as it was the highest performing model in the original paper. These models were trained on an 80:20 split of the NDCLD'15 dataset -- 5,840 images in the training set and 1,460 images in the validation set. The models were ranked by their performance on the validation set and then added one at a time to an ensemble, as can be seen in Fig. \ref{fig:nd_validation}. The highest validation performance was achieved with a 10 model ensemble, which gave a CCR of 99.86\%. 
\hfill

\begin{figure}[!htb]
    \centering
    \includegraphics[width=0.5\textwidth, keepaspectratio]{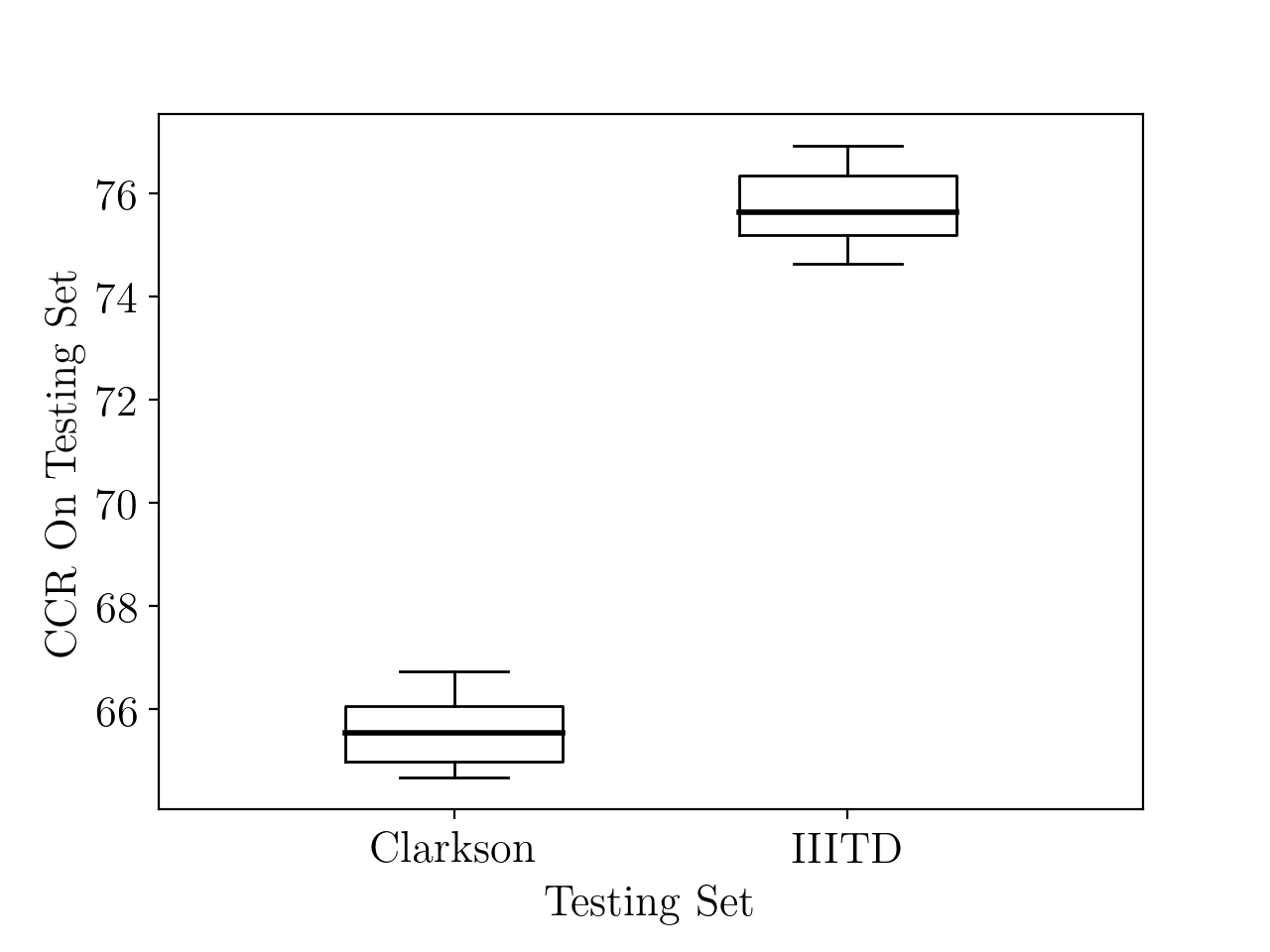}
    \caption{Box plots presenting the correct classification rates seen when validation is performed on one dataset (ND) and testing on others. Bold bars denote median values, height of each boxes equals to an inter-quartile range (IQR) spanning from the first (Q1) to the third (Q3) quartile, and the whiskers span from Q1-1.5*IQR to Q3+1.5*IQR.}
    \label{fig:prelim_results}
\end{figure}

This ensemble, trained on NDCLD'15 dataset, was then tested in a {\bf cross-dataset} scenario to assess how the proposed benchmark generalizes to unknown data. We decided to use Clarkson and IIITD training partitions from the LivDet-Iris 2017 benchmark for that purpose. To estimate a variance of test results, ten testing iterations were performed, with each iteration consisting of a randomly selected set of images that was half the size of the overall test set. As can be seen in Fig. \ref{fig:prelim_results}, the ensemble trained on NDCLD'15 dataset does not produce results that are on par with the LiveDet-Iris 2017 winner, what suggests that the generalization capabilities of a solution achieving an excellent CCR=99.8\% in same-dataset scenario is limited. Therefore, additional BSIF filters and additional classifiers were considered in the extended case to improve the poor cross-dataset performance of the base case.

\subsection{Extended Case}
A final ensemble of models, made available with this paper, was trained on the entire NDCLD'15 dataset.  For each combination of $s \in \{3, 5, 6, 7, 9, 10, 11, 13, 14, 15, 17, 18, 22, 26, 30, 34\}$ and $n \in \{5, 6, 7, 8, 9, 10, 11, 12\}$, three models (SVM, RF, and MLP) were trained on features extracted from the NDCLD'15 dataset, giving 360 total models available in the ensemble.

To select the strongest models, either the Clarkson or IIITD dataset was taken as the validation set. The other dataset was excluded from the validation procedure to be used as the testing set. Each model was tested on the validation set and then ranked by the correct classification rate measured on the validation set. The models were then added one by one from strongest to weakest to a majority voting test on the validation set to determine the optimal number of models to use in majority voting. For example, if a peak was found when using the top eight models for majority voting on the validation set, these top eight models were selected to test for ensemble performance on the testing set. This procedure ensures that the other dataset will be {\bf unknown} to the algorithm and the results will be a true indicator of {\bf cross-dataset} performance.

\begin{figure*}[!htb]
   
    \centering
    \begin{subfigure}[t]{\textwidth}
        \includegraphics[width=\textwidth, keepaspectratio]{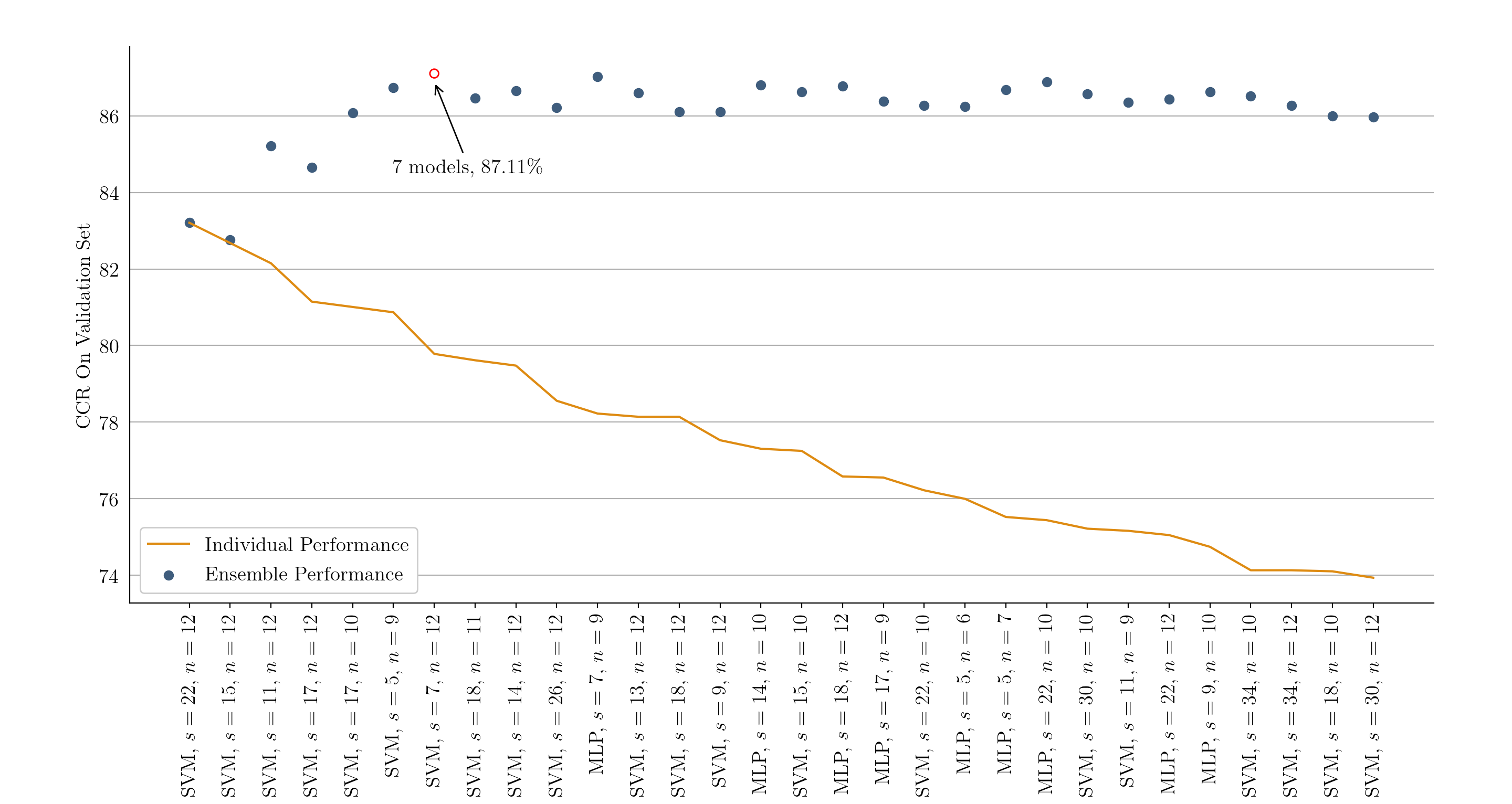}
        \caption{Building a classification ensemble on {\bf Clarkson} dataset.}
    \end{subfigure}\hfill
    \begin{subfigure}[t]{\textwidth}
        \includegraphics[width=\textwidth, keepaspectratio]{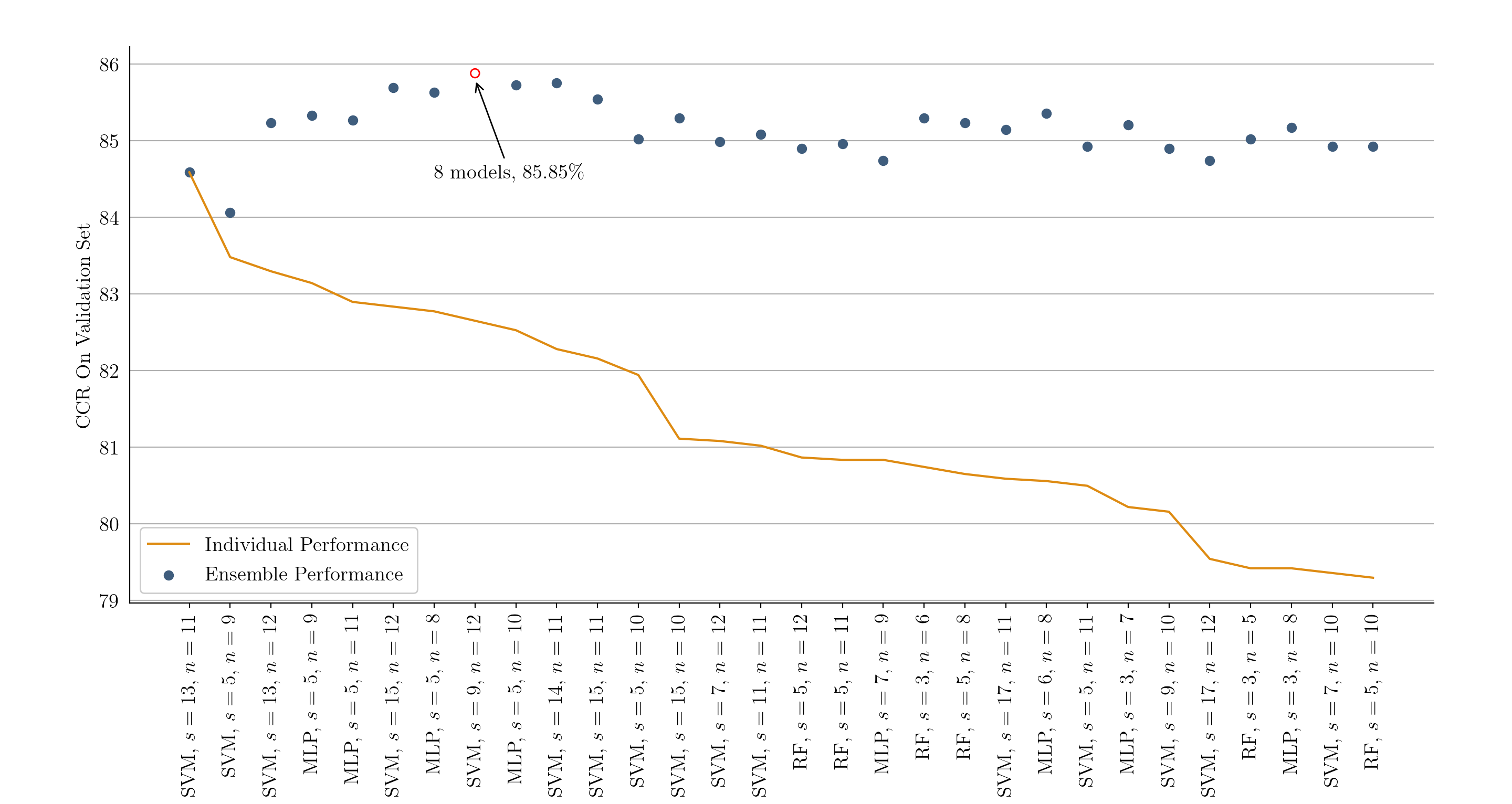}
        \caption{Building a classification ensemble on {\bf IIITD} dataset.}
    \end{subfigure}\hfill
    
    \caption{Correct classification rate measured on validation sets when the top ranking models were added to an ensemble of classifiers one at a time. For Clarkson (a), the maximum performance was achieved when 7 models were used in the ensemble, giving a CCR of $87.11\%$. For IIITD (b), the maximum performance was achieved when 8 models were used in the ensemble, giving a CCR of $85.85\%$.}
     \label{fig:validationResults}
    
\end{figure*}  

The results from validation on Clarkson and IIITD can be seen in Fig. \ref{fig:validationResults} where both the individual model performance and the ensemble performance are shown. For Clarkson, the highest performance (determined by the correct classification rate) comes when seven models are used in majority voting, giving a CCR of 87.11\%. For IIITD, the highest performance comes when eight models are used in majority voting, giving a CCR of 85.88\%. These two best-performing ensembles are passed to the testing phase.

%

%

\subsection{Cross-Dataset Testing}

\begin{figure}[!htb]
    \centering
    \includegraphics[width=0.5\textwidth, keepaspectratio]{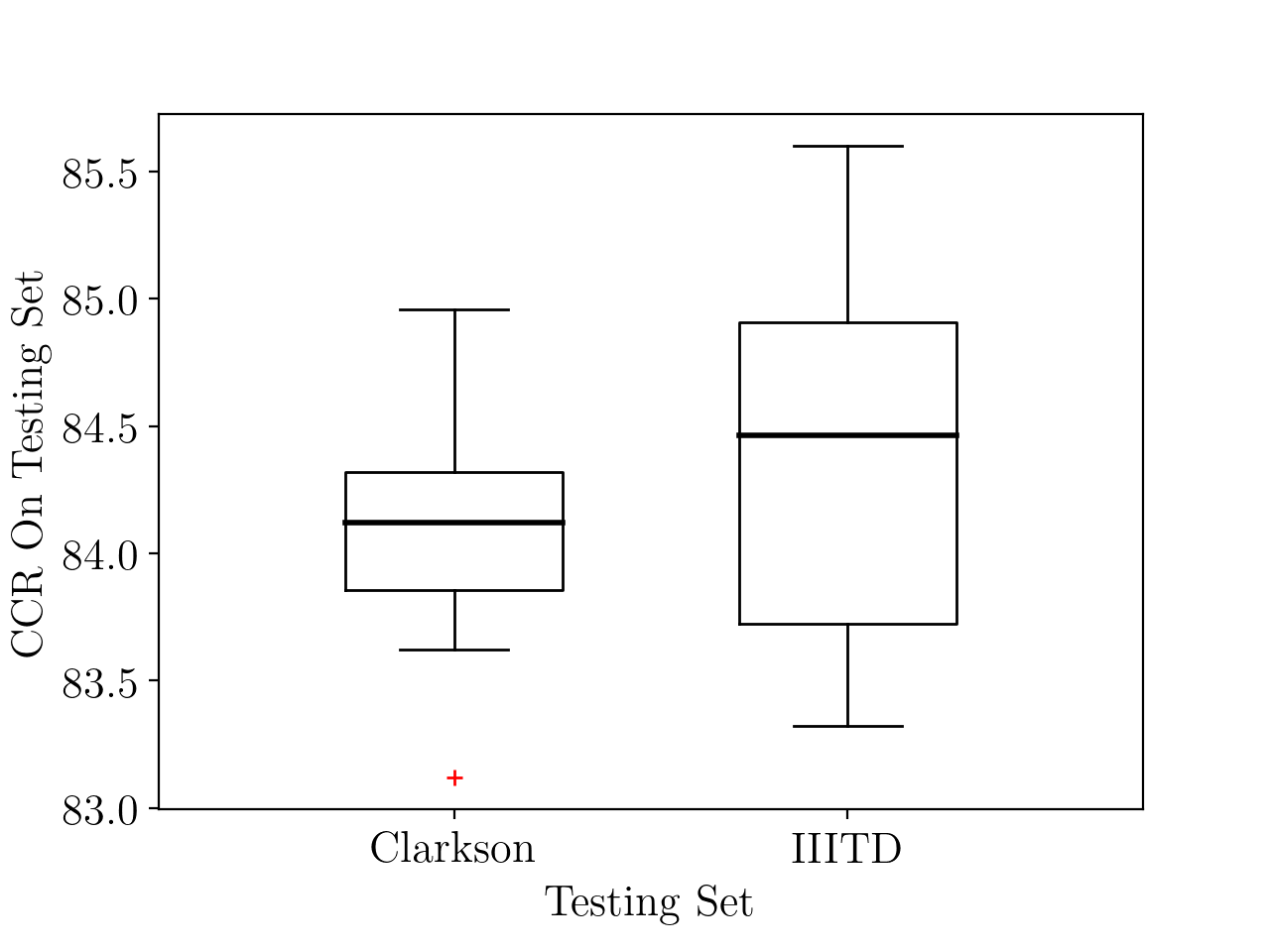}
    \caption{Box plots presenting the correct classification rates seen when validation is performed on one dataset and testing on the other. Bold bars denote median values, height of each boxes equals to an inter-quartile range (IQR) spanning from the first (Q1) to the third (Q3) quartile, and the whiskers span from Q1-1.5*IQR to Q3+1.5*IQR.}
    \label{fig:test_results}
\end{figure}

After the ensemble of models was selected on the validation set, it was used to classify the other dataset, which served as the testing set. To estimate a variance of test results, ten testing iterations were performed, with each iteration consisting of a randomly selected set of images that was half the size of the overall test set. The results of this test can be seen in Fig. \ref{fig:test_results}. The 7 models selected using Clarkson as the validation set were able to achieve a median correct classification rate of $84.45\%$ on IIITD, and the 8 models selected using IIITD were able to achieve a median correct classification rate of $84.11\%$ on Clarkson. These results are close to the performance achieved by the LivDet-Iris 2017 winner (90\% on Clarkson dataset and 83\% on IIITD dataset). Such comparison should be done with care, as the LivDet-Iris 2017 paper reports combined results of textured contact lens {\bf and} printouts detection. However, if iris printouts are -- on average -- easier to detect than textured contact lenses, as suggested by the LivDet-Iris 2017 results, the obtained accuracy for this open source benchmark compares favorably to the LivDet winner. These results show that a larger ensemble can be more robust to cross-dataset tests.

\subsection{Models for Release}

To provide a complete solution in the open source release, the 360 models (120 each of SVM, RF, and MLP) that were trained using the NDCLD'15 database have been included as a ready-to-use solution for iris PAD. They can be used in the ensembles selected from validation on Clarkson or IIITD, or from ensembles selected from performance on a novel validation set provided by the user.

\section{Summary}
\label{sec:conclusions}

This paper offers the first, known to us, open-source software solution for iris presentation attack detection.
It is based on a recent and effective methodology of using Binary Statistical Image Features and an ensemble of classifiers to detect textured contact lenses. A trained ensemble of classifiers, added to this initial version, achieves a correct classification rate of 84\% on challenging cross-dataset tests, in a close-set scenario and with only best guess iris segmentation required. This result is similar to the cross-dataset classification accuracy achieved by the most recent LivDet-Iris competition winner, which makes this implementation a useful benchmark for iris PAD.

This software allows for retraining the ensemble with other datasets, defining which classifiers form the ensemble, and calculating BSIF-based features that can be used to test other classifiers worth adding to the ensemble. The long-term goal of this effort is to build an open-source baseline methodology for iris PAD, for instance for the next editions of the LivDet-Iris competitions, starting from a recent and effective algorithm of textured contact lens detection.

{\small
\bibliographystyle{ieee}
\bibliography{ref}
}

\end{document}